\documentclass[10pt,twocolumn,letterpaper]{article}

\usepackage{wacv}
\usepackage{times}
\usepackage{epsfig}
\usepackage{graphicx}
\usepackage{amsmath}
\usepackage{amssymb}
\usepackage{verbatim}
\usepackage{multirow}

\usepackage{caption}
\usepackage{subcaption}

\newcommand{\mr}[1]{\mathrm{#1}}
\DeclareMathOperator*{\argmax}{arg\,max}

\DeclareMathOperator{\E}{\mathbb{E}}
\newcommand{\Y}{\widetilde{Y}}
\newcommand{\ppm}{$\,\pm\,$}
\newcommand{\MI}{\mathcal{I}}

\usepackage[pagebackref=true,breaklinks=true,letterpaper=true,colorlinks,bookmarks=false]{hyperref}

\wacvfinalcopy 

\def\wacvPaperID{1281} 

\ifwacvfinal\fi
\begin{document}

\title{Transformation-based Regularization for Information-based Deep Clustering}

\author{Jizong Peng \hspace{1cm} Christian Desrosiers  \hspace{1cm} Marco Pedersoli\\
ETS Montreal \\
{\tt\small jizong.peng.1@etsmtl.net}
}

\maketitle
\ifwacvfinal\thispagestyle{empty}\fi

\begin{abstract}
The joint learning of a feature representation and a clustering of the training data is a powerful approach for unsupervised learning in which data is automatically divided in clusters that represent semantic classes.
Recent clustering approaches based on mutual information maximization have achieved excellent results, sometimes comparable with fully-supervised approaches.
In this work, we present a generalization of information-based deep clustering where two key factors are evaluated: i) the variables for which we want to maximize the mutual information, ii) the regularization of the mutual information loss by the use of image transformations. 
Through an extensive analysis, we show that maximizing the mutual information between a sample and its transformed version, with an additional regularization 
to make the learning smoother, outperforms previous approaches and leads to state of the art results on three different datasets. 
Additional experiments show that the proposed method largely outperforms disentangling methods for classification tasks and is useful as unsupervised initialization for supervised learning.

\end{abstract}


\section{Introduction}

In deep learning, supervised methods have shown excellent performance, sometimes even surpassing human level \cite{krizhevsky12imagenet,he2015delving}. However, these methods require large datasets with fully annotated data, which cannot be afforded in many cases \cite{vondrick13efficiently}. 
Instead, unsupervised methods can learn from data without annotations, which is very appealing given the large amount of data that can easily be collected from various sources such as social media \cite{schroff2011harvesting}.
In this paper, we are interested in the unsupervised learning problem of deep clustering, which consists in learning to group data into clusters, while at the same time finding the representation that best explains the data. Jointly learning to group data (clustering) and represent data (representation learning) is an ill-posed problem which can lead to poor or degenerate solutions \cite{yang16joint,xie2016unsupervised,bojanowski18deep}. 
A principled way to avoid most of these problems is mutual information
\cite{paninski2003estimation}. Mutual information is a powerful approach for clustering because it does not make assumptions about the data distribution 
and reduces the problems of mode collapse, where most of the data is grouped in a single large cluster \cite{bojanowski18deep}.


In recent publications, two papers obtained outstanding results for deep clustering by using mutual information in different ways. The first one, IMSAT \cite{hu2018imsat} maximizes the mutual information between input data and the cluster assignment, and regularizes it with virtual adversarial samples \cite{miyato2019virtual} by imposing that the original sample and the adversarial sample should have similar cluster assignment probability distribution (by minimizing their KL divergence). The second one, IIC \cite{ji2018iic}, maximizes the mutual information of the cluster assignment between a sample and the same sample after applying a geometrical transformation and no additional regularisation is used.

In this paper, we aim to analyze and better understand these algorithms by decomposing them in two basic building blocks: the information-based loss and a regularization term based on image transformations.
We build a generalization of information-based clustering approaches in which IMSAT and IIC are special cases. We consider the two different ways to use mutual information for clustering and we evaluate the effect of a regularization based on different image transformations currently used for data augmentation such as Geometrical transformations, Gaussian noise, Virtual Adversarial Training (VAT)~\cite{miyato2019virtual}, Mixup~\cite{zhang2017mixup}, Cutout~\cite{devries17cutout}. 
Form our extensive evaluation on three different datasets we found that:
\emph{i}) maximizing the mutual information between a cluster distribution and its transformed version seems to be more robust than other approaches when dealing with challenging datasets;
\emph{ii}) adding a transformation-based regularization to the mutual information loss can  make the training smoother and leads to better clusters;
\emph{iii}) using geometrical transformations for the mutual information loss together with a regularization based on VAT \cite{miyato2019virtual} leads to improved results in most of the datasets.
Additionally, our best model outperforms popular methods for disentangling representations and can be used as initialization to improve supervised training.

In the reminder of this paper, we first introduce related work in section \ref{sec:related}, with special attention to the two mentioned methods. Then, in section \ref{sec:protocol}, we propose the experimental protocol by defining the different components of our experiments. Finally, we report results in section \ref{sec:experiments} and draw conclusions in section \ref{sec:discussion}. 

\section{Related work}
\label{sec:related}

\paragraph{Mutual information}
Mutual information $\MI(X,Y)$ is a information-theoretic criterion to measure the dependency between two random variables $X, Y$ \cite{NIPS1991_440}. It is defined as the KL divergence between the joint distribution $p(X,Y)$ of two variables and the product of their marginal: $\MI(X;Y) = D_{\mr{KL}}(p(X,Y),p(X)p(Y))$. 
The criterion of maximizing mutual information for clustering is first introduced in \cite{NIPS1991_440}, as the \textit{firm but fair} criterion. In this case the mutual information between {input data} (i.e, image or representation) and {output categorical distribution} is maximized, believing that the class distribution can be deduced given the input. This principle is extended in \cite{krause10discr}, in which mutual information is maximized with additionally an explicit regularization, such as $L_{2}$ loss. This helps to avoid too complex decision boundaries. 

In \cite{viola1997alignment,pluim2003mutual,thevenaz2000optimization}, mutual information is used to align inputs with different modalities \cite{viola1997alignment,pluim2003mutual,thevenaz2000optimization}, since with different modalities normal distances are meaningless. Finally, mutual information is also used as regularization in a semi-supervised setting \cite{manohar2015semi}.
%
Recently, DeepInfoMax \cite{hjelm2018learning} simultaneously estimates and maximizes mutual information between {images} and learns {high-level representations}. However, estimating the mutual information of images is challenging and requires complex techniques \cite{belghazi2018mine}. 
Finally, two recent techniques for deep clustering based on mutual information are 
IMSAT \cite{hu2018imsat} and IIC \cite{ji2018invariant}. As these are the starting point of this study, they will be analyzed in more detail in section \ref{sec:protocol}.
\noindent\textbf{Self-supervised approaches}
Self-supervised learning has recently emerged as a way to learn a representative knowledge based on non-annotated data. 
The main principle is to transfer the unsupervised task to a supervised one by defining some \textit{pseudo labels} that are automatically generated by a \textit{pretext task} without involving any human annotations \cite{jing2019self}. The network trained with the pretext task is then used as the initialization of some downstream tasks, such as image classification, semantic segmentation and object detection. It has been shown that a good  pretext task can help improve the performance of the downstream task \cite{pathak2017curiosity, doersch2017multi, noroozi2018boosting, caron2018deep,ahsan2019video}.
Without the access to label information, the pretext task is usually defined based on the data structure and is somewhat related but different from the downstream tasks. Various pretexts have been proposed and investigated. Similarity-based methods \cite{noroozi2018boosting,caron2018deep} design a pretext task that let the network learn the semantic similarities between image patches. Likewise, \cite{noroozi2016unsupervised,ahsan2019video,wei2019iterative} trained a network to recognize the spatial relationship between image patches. See \cite{jing2019self} for a complete survey on such methods.

In this work, we argue that if the task we want to learn with non-annotated data is classification then the best pretext task is clustering. With clustering, the pretext task is very close to the downstream task of classification. In fact, clustering aims to group the data in a meaningful way and therefore split the data into categories. If these categories are not only visually similar but also semantically, classification and clustering become the same task. In other words, with clustering, there is not need for a downstream task. By assigning the most likely category to each cluster, our clustering method becomes a classifier. In the experimental evaluation, we will compare the performance of our information based clustering with state-of-the-art self-supervised learning approaches. 
For doing that, we use two simple assumptions: i) the sample distribution per class is known (normally uniform) and ii) the exact number of classes, which corresponds to the number of clusters is also known.

%
%
%
\noindent\textbf{Clustering approaches}
Clustering has been long time studied before the deep learning era. K-means 
\cite{kanungo2002efficient} and GMM algorithms \cite{banfield1993model} were popular choices given representative features. Recently, much progress has been made by jointly training a network that perform feature extraction together with a clustering loss \cite{xie2016unsupervised,li2018discriminatively,ghasedi2017deep, caron2018deep}. 
Deep Embedded Clustering (DEC) \cite{xie2016unsupervised} is a representative method that uses an auto-encoder as the network architecture and a cluster-assignment hardening loss for regularization. Li et al.  \cite{li2018discriminatively} proposed a similar network architecture but with a boosted discrimination module to gradually enforce cluster purity. DEPICT \cite{ghasedi2017deep} improved the clustering algorithm's scalability by explicitly leveraging class distributions as prior information. 
DeepCluster \cite{caron2018deep} is an end-to-end algorithm that jointly trains a Convnet with K-means and groups high-level features to $N$ pseudo labels. Those pseudo labels are in turn used to retrain the network after each iteration. 
In this work, 
we focus on two state of the art information-based clustering approaches \cite{hu2018imsat,ji2018invariant} and analyze their different components and how they can be combined in a meaningful way. 

\section{Information-based clustering}
\label{sec:protocol}
We consider information-based clustering as a family of methods having two main components: the mutual information loss and possibly a regularisation based on image transformations. The maximization of the mutual information aims to produce meaningful groups of data, i.e. clusters with a similar representation and with a even number of samples. On the other hand, transformations are used to make the learned representation locally smooth and ease the optimization in a similar way as in data augmentation. For these two components, we consider and evaluate different possible choices and their combination.

\subsection{Mutual Information losses}
\label{sec:mivar}
\noindent\textbf{$\bold{I(Y,\Y)}$:}
This formulation introduced in IIC \cite{ji2018invariant} maximizes the mutual information between the clustering assignment variable $Y$ and the clustering assignment of a transformed sample $\Y$. 
Mutual information is defined as the KL divergence between the joint probability of two variables and the product of their marginals \cite{NIPS1991_440}:
\begin{equation}
    \MI(X_1,X_2) = D_{\mr{KL}}\big(p(X_1,X_2), \, p(X_1)\,p(X_2)\big).
    \label{equ:MIdefinition}
\end{equation}
It represents a measure of the information between the two variables. If the variables are independent, the mutual information is zero because the joint will be equal to the product of the marginals.
Thus, we need to estimate the joint probability of the clustering assignment and its transformed version $p(Y,\Y)$, as well as the marginals $p(Y)$ and $p(\Y)$.
The marginals are defined as
\begin{equation}
    p(Y) = \E_X[p(\,Y|X)\,], \ \ \ p(\Y) = \E_X[\,p(Y|T(X))\,],
\end{equation}
and can be empirically estimated by averaging output mini-batches.
For the joint, we compute the dot product between $Y|X$ and $Y|T(X)$ for each sample and marginalize over X:
\begin{equation}
    p(Y,\Y) \ = \E_X[p(Y|X) \, \cdot \, p(Y|T(X))^\top].
\end{equation}
For each sample, the joint probability of $c_1 \in Y|X$ and $c_2 \in Y|T(X)$ is $c_1\,c_2$. 
For a single sample, by construction the joint and the marginals will be equivalent. However, when marginalizing the joint over samples $X$, the final $p(Y,\Y)$ will be different than $p(Y)p(\Y)$. 


This formulation maximizes the predictability of a variable given the other. It is different than enforcing KL divergence between two distributions because:
\emph{i}) it does not enforce the two distributions to be the same, but only to contain the same information. For instance, one distribution can be transformed by an invertible operation without altering the mutual information.  
\emph{ii}) it penalizes distributions that do not have uniform marginal, i.e. all the cluster should contain an even number of samples.\\[2mm]
%
\noindent\textbf{$\bold{I(X,Y)}$:}
We consider the MI formulation used as loss in IMSAT \cite{hu2018imsat}. It connects the input image distribution $X$ with the output cluster assignment of the used neural network $Y$ and is the most common way to use mutual information for clustering \cite{NIPS1991_440,krause10discr}.
As the input of the neural network is a continuous vector, estimating its probability distribution is hard and we cannot use directly equ.~(\ref{equ:MIdefinition}). Instead, in IMSAT the mutual information between input and output is computed as: 
\begin{multline}
    \MI(X,Y) \ = \ H(Y) - H(Y|X) \\
        \ = \E_Y[\,\log \E_X[\,p(Y|X)\,]\,] - \E_{X,Y}[\,\log p(Y|X)\,]
\end{multline}
In this formulation, the mutual information is easy to compute because it is the difference between the entropy of the output $H(Y)$ and the conditional entropy $H(Y|X)$. Both quantities can be approximated in a mini-batch stochastic gradient descent setting. $H(Y)$ is approximated as the entropy of the average probability distribution $p(Y|X)$ over the given samples, while $H(Y|X)$ is approximated as average of the conditional entropy of each sample. 
As entropy is a non-linear operation, the two quantities are different. $H(Y)$ is maximized when the probability of each cluster is the same, i.e. the output has the same probability distribution for each cluster. On the other hand, $H(Y|X)$ is minimized when, for each sample $X$, $p(Y|X)$ has most of the probability distribution assigned to a single cluster, i.e. the model is certain about a given choice. Combining the two imposes that the clustering chooses a single cluster for every sample and, globally, each cluster contains the same number of samples. In case the class distribution is not uniform, 
another distribution $C$ can be enforced by $D_{\mr{KL}}(p(X), p(C))$. In our experiments, we limit ourselves to uniform class distribution.
Notice that, in IMSAT, the authors add an hyper-parameter MI formulation $\lambda H(Y)-H(Y|X)$.  This parameter is the minimum value that ensures that the data is evenly distributed on all clusters. We follow the same approach as in the original paper.\\[2mm]
\subsection{Regularization:}
In the context of this work, we call regularization the additional loss that penalizes when the output of the model for the original image and the transformed image are different.
Although in unsupervised settings there is no real difference between loss and regularization because in both cases the training is performed without annotations, here we consider this KL divergence as a regularization because it cannot be used alone for clustering. Another term such as $H(Y)$ is required to enforce an even distribution of samples in the clusters. 

As we have access to the clustering probability distribution $p(Y|X)$ of a sample $X$ and its transformed version $p(Y|T(X))$, we use the KL divergence as penalty term $D_{\mr{KL}}(p(Y|X),p(Y|T(X)))$.
While for $\MI(X,Y)$ this regularization is fundamental for good results \cite{krause10discr,hu2018imsat}, for $\MI(Y,\Y)$ the original paper did not use any additional regularization.
In this work, we analyze the effect of regularization with different transformations for both approaches. The following section describes in more detail the used transformations.


\subsection{Transformations}
\label{sec:transf}
Transformations seem to be a key component of MI-based clustering. To be useful, any transformation needs to change the appearance of the image (in terms of pixels) while maintaining its semantic content, i.e. the class of the image. We can think of these transformations as a pseudo ground-truth that helps to train the model. 
In this work, we consider five types of transformations: Geometrical, Gaussian, Adversarial, Mixup and Cutout.\\[2mm]
\noindent\textbf{Geometrical:}
Geometrical transformations are the image transformations that are normally used for data augmentation. 
As in \cite{ji2018iic}, we use random crop, resize at multiple scales, horizontal flip, and color jitter. 
Note that some transformations can actually change the category of a class. For instance, on MNIST \cite{lecun1998gradient} a dataset composed of numbers, a crop of a \textbf{$6$} can zoom in the lower circle and look very similar to a $0$. We will further discuss this problem in the experimental results. 
\\[2mm]
\noindent\textbf{Gaussian:}
Adding Gaussian noise to the input space, as a regularization method, has been proposed in \cite{rifai2011adding}. It has been demonstrated that noise injection can effectively reduce the L2-norm of the Jacobian's mapping function with respect to the input \cite{rifai2011adding}, thus implicitly enforce regularization similar to L2 constrains. In this paper, we consider to add Gaussian independent and identically distributied noise having 0 mean and $\theta_{\mathrm{std}}$ standard derivation to each pixel of the three channels of the original input image $x$: 
	\begin{equation}
		x_{\mr{Gauss}} = \ x + \mathcal{N}(0,\theta_{\mr{std}}). \\
	\end{equation}
\\[2mm]
\noindent\textbf{Adversarial:}
Adversarial samples \cite{yuan2019adversarial} are samples that are slightly modified by an adversarial noise which is usually unnoticeable by the human eye, but can induce a neural network to misclassify an example.
Recently, methods based on adversarial examples have attracted a lot of attention because they can easily fool machine learning algorithms and thus represent a threat to any system using machine learning \cite{su2019one,chou2018sentinet}. It has been shown \cite{madry2017towards} that adding those samples during training can help to improve the robustness of the classifier. In this study, we use Virtual Adversarial Training(VAT) \cite{miyato2019virtual}, an extension of adversarial attack that can also be used for non-labelled samples and has shown promising results for fully-supervised, semi-supervised \cite{miyato2019virtual} and unsupervised learning \cite{hu2018imsat}.
The adversarial noise $r$ can be found as the value within a certain neighbourhood $\epsilon$ that maximizes the distance $D$ between the probability distribution of the original sample $x$ and the transformed sample $x+r$: 
	\begin{equation}
		r_{\mr{vadv}} \ = \ \argmax_{\|r\|_{2}\,\leq\,\epsilon} \ D\big[p(y|x,\theta), \,p(y|x+r)\big],
		\label{equ:VAT}
	\end{equation}
where $D[p_{1}(.), p_{2}(.)]$ is a divergence function, usually defined as KL-divergence. In practice, equ.~(\ref{equ:VAT}) can be optimized in order to find $r$ with a few iterations of the power method \cite{journee2010generalized}.
Note that we could also experiment with adversarial geometrical transformations as in  \cite{peng2018jointly}, but we leave this direction as future work.\\[2mm]
\noindent\textbf{Mixup:}
This is a simple data augmentation technique that has proven successful for supervised learning \cite{zhang2017mixup}. It consists on creating a new sample $x_{\mr{new}}$ and label $y_{\mr{new}}$ by linearly combining two training samples $x_1$ and $x_2$ (e.g. images) and labels $y_1$ and $y_2$ (e.g. class probabilities):
	\begin{equation}
	\begin{split}
		x_{\mr{new}} \ = \ \alpha x_1 + (1-\alpha) x_2 \\
		y_{\mr{new}} \ = \ \alpha y_1 + (1-\alpha) y_2.
	\end{split}
	\end{equation}
$\alpha$ is the mixing coefficient and is normally sampled from a $\beta$ distribution.
Although very simple and effective, Mixup has received multiple criticisms because it is clear that the generated images do not represent real samples. However, the $\beta$ distribution has most of its mass near 0 and 1, which means that in most of the cases the mixed samples look very similar to one of the samples, but with a structured noise coming from the other image.
This transformation differs from the previous one because it requires two input samples to generate a new one. Thus, to use it in our family of algorithms, we had to adapt it.
For $\MI(Y,\Y)$, as before we consider $Y=\E_X[p(Y|X)]$ the expected output of real samples $X$, while $\Y=\E_{X,X_2,\alpha}[\alpha p(Y|X) (1-\alpha)P(Y|X_2)]$ is now the output associated to mixup samples generated using the same real samples $X$ in combination with other samples $X_2$ that is randomly selected. 
Mixup can also be used as direct regularization (see next section). In this case, the first output distribution is associated to real samples while the second is associated to mixup samples built as above.
\\[2mm]
\noindent\textbf{Cutout:}
Cutout \cite{devries17cutout} is an effective data augmentation method to boost the generalization of a neural network, and has been used in { (self)-supervised learning \cite{pathak2016context}, and weakly supervised localization \cite{singh2017hide}}. Cutout forces the network to consider the global information of an image by randomly masking some regions of it. We found this data augmentation suitable for clustering, in the sense that a masked image should have very similar cluster category as the original image. 

\section{Experimental Setup}
\label{sec:experiments}
Our main experiment evaluates on the three datasets (presented below) the two identified components for information based clustering: information based losses and regularization based on image transformations.
For completeness, we have reported all results with all combinations of the different components in the supplementary material. 

\subsection{Datasets}
We evaluate the different methods on 3 datasets:
\begin{itemize}
\item MNIST dataset \cite{lecun1998gradient} of hand-written digit classification consists of 60,000 training images and 10,000 validation images. 10 classes are evenly distributed in both train and test sets. Following common practice, we mix the training and test set to form a large training set. During training, we do not show any ground truth information, while for testing, we use image annotations to find a mapping between true class label and cluster assignment, thus assessing the clustering performance by the classification accuracy.
\item CIFAR10 \cite{krizhevsky2009learning} is a popular dataset consisting of 60,000 32$\times$32 color images in 10 classes, with 6,000 images per class. Similar to MNIST dataset, we mix the 50,000 training images with 10,000 test images to build a larger dataset for clustering.
\item SVHN \cite{netzer2011reading} is a real-world image dataset for digit recognition, consisting of 73,257 digits for training, 26,032 digits for testing. Images come from natural scene images. We adopt the previously described strategy to use this dataset too.
\end{itemize}
\subsection{Evaluation Metric}
Our method groups samples into clusters. If the grouping is meaningful, it should be related to the datsest classes. Thus, in most of our experiments, we use classification accuracy as measure of the clustering quality. This makes sense because the final aim of this approach is exactly to produce a classifier without using training labels.
This accuracy is based on the best possible one-to-one mapping (using the Hungarian method \cite{Kuhn55thehungarian}) between clustering assignment and ground truth label (assuming they share the same number of classes). 
We run the experiments 3 times with different initialization and report mean and standard deviation values. 

\subsection{Implementation Details}
In order to provide a fair comparison, we use the same network for a given dataset across methods. For both MNIST and SVHN dataset, we borrow the setting of IIC \cite{ji2018iic}, using a VGG-based convolutional network as our backbone network. For CIFAR-10, we use a ResNet-34 \cite{he2016deep} based network. It is worth mentioning that, in original IMSAT paper \cite{hu2018imsat}, the used network was just a 2 fully connected layers with pre-trained features on CIFAR-10 or GIST features \cite{oliva2001modeling} on SVHN. Instead, in this work, we want to compare all results on the same convolutional architecture and without pre-trained models or any hand-crafted features. 

For our best method, as in \cite{ji2018iic} we use two additional procedures to further improve results. The first one, over-clustering, consist in using more clusters than the number of classes in the training data. This can help to find sub-classes and therefore reduce the intra-class variability on each cluster. The second consists in splitting the last layer of the network in multiple final layers (there called heads) and therefore multiple clusters. This can increase diversity and acts as a simplified form of ensembling. Combining these two techniques can highly boost the final performance of the clustering approach. However, they also increase the computational cost of the model. Thus, for the evaluation of all configurations in a same setting, we use a basic model without additional over-clustering or multiple final layers. However, for our best configuration, we retrained it with 5 final layers with 10 clusters (as the number of classes) and other 5 final layers with 50 clusters for MNIST and SVHN or 70 clusters for CIFAR10.



\begin{table*}[ht]
\begin{center}
\begin{small}
\begin{tabular}{l| c|c|c}
\textbf{Method} & MNIST & CIFAR10 & SVHN\\
\hline
$\MI(X,Y)$&  42.6\ppm3.9  & 16.6\ppm1.0 &  15.4\ppm1.5  \\
$\MI(X,Y)\,+\,D_{\mr{KL}}\big(p(Y|X),p(Y|\text{Geo}(X))\big) $ & $43.8\pm47.8$ & $14.5\pm0.7$ & $17.4\pm0.8$ \\
$\MI(X,Y)\,+\,D_{\mr{KL}}\big(p(Y|X),p(Y|\text{VAT}(X))\big) $  & $\textbf{97.7}\pm0.3$ & $18.4\pm0.4$ & $\textbf{17.4}\pm1.2$ \\
$\MI(X,Y)\,+\,D_{\mr{KL}}\big(p(Y|X),p(Y|\text{Mixup}(X))\big) $ & $34.4\pm12.0$ & $\textbf{20.1}\pm1.2$ & $15.5\pm1.3$ \\
$\MI(X,Y)\,+\,D_{\mr{KL}}\big(p(Y|X),p(Y|\text{Cuout}(X))\big) $ & $60.7\pm9.0$ & $18.1\pm2.2$ & $16.2\pm1.5$ \\
$\MI(X,Y)\,+\,D_{\mr{KL}}\big(p(Y|X),p(Y|\text{Gauss}(X))\big) $ & $65.8\pm12.0$ & $17.2\pm1.3$ & $13.4\pm0.4$ \\
\hline
$\MI(Y,\Y)$ & {\textbf{97.9}\ppm0.0} & 31.9\ppm1.1 & 28.0\ppm4.0   \\
$\MI(Y,\Y)\,+\,D_{\mr{KL}}\big(p(Y|X),p(Y|\text{VAT}(X))\big)$  &  $97.6\pm0.1$ & $37.5\pm2.9$ & $34.3\pm3.9$ \\
$\MI(Y,\Y)\,+\,D_{\mr{KL}}\big(p(Y|X),p(Y|\text{Mixup}(X))\big)$  & $94.1\pm6.7$ & $\textbf{39.2}\pm0.9$ & $27.0\pm0.8$ \\
$\MI(Y,\Y)\,+\,D_{\mr{KL}}\big(p(Y|X),p(Y|\text{Cutout}(X))\big)$  & $97.9\pm1.3$ & $33.9\pm3.7$ & $\textbf{39.6}\pm2.9$ \\
$\MI(Y,\Y)\,+\,D_{\mr{KL}}\big(p(Y|X),p(Y|\text{Gauss}(X))\big)$  & $94.4\pm2.5$ & $29.7\pm0.9$ & $35.2\pm3.3$ \\
\end{tabular}%
\end{small}
\end{center}
\caption{\textbf{Transformation-based Regularization}. We consider the information-based losses presented in section \ref{sec:mivar} and report results on the three datasets validation sets with different regularisation transformations.} 
\label{tab:losses}
\end{table*}

\section{Experiments}
In this section we evaluate the different components that we have considered for clustering. First of all, in the next subsection we evaluate the clustering performance of the two mutual information losses. Next, we report the effect of adding a regularization term with different image transformations to the analyzed losses. Then, we consider the effect of using different transformations in the computation of the mutual information between an image and its transformed version. 
Finally, we show that the proposed clustering can be used to initialize the parameters of a network to improve image classification.
We compare our best model with several well-known methods for disentangling representation on the task of linear supervised classification using the representation learned by the respective methods.

\subsection{Mutual Information Loss}
{Table~\ref{tab:losses} report results on the two ways of using mutual information for clustering as explained in section \ref{sec:mivar}.  $\MI(X,Y)$ considers the loss between input image and the output cluster. While this form of using the entropy is the most commonly used for clustering data \cite{NIPS1991_440,krause10discr}, we notice a large gap in performance compared to $\MI(Y,\Y)$. This is probably because the latter already includes in the mutual information optimization a (geometrical) transformation, which brings more prior knowledge on the problem.
In the next subsection we analyze the same two methods when explicitly adding an additional regularization term.

\subsection{Transformation-based Regularization}
In this section we consider the performance of the two information based losses when adding regularization based on the different transformations described in section \ref{sec:transf}.
For each loss and each dataset we report in bold the transformation that leads to the best accuracy.
For both losses $\MI(X,Y)$ and $\MI(Y,\Y)$, adding a regularisation term based on $D_{KL}$ seems to help. 

For $\MI(Y,\Y)$ in the table we consider $\Y$ as the marginal probabilities for samples with geometrical transformations. 

Even though there is not a clear winner, a regularization based on VAT seems to lead to high performance on all datasets. 
For the following experiments we then use a model that maximises the mutual information between and example and its geometrical transformation and an additional regularization based on VAT.


In Fig.~\ref{fig:transform} we visually compare the clustering performance of different mutual information losses with or without regularization. 
Each row, represent a cluster found in an unsupervised way. If the samples in each row belong to the same class, the clustering has managed to find a meaningful grouping strategy and the associated classification accuracy will be high. From visual inspection and similarly to the accuracies in table~\ref{tab:losses}, we observe that the clusters obtained with regularization have better similarity with semantic classes. 

\subsection{Transformations on the Mutual Information}
In the previous section we have seen that maximizing the mutual information between and image and its transformed version $\MI(Y,\Y)$ leads to better performance than maximizing the mutual information between input and output $\MI(X,Y)$ as in \cite{hu2018imsat}.
Therefore here we focus on the first method and evaluate its performance when changing the transformations on $\Y$.
In Table \ref{tab:transf} we see that when applying transformations directly on the mutual information loss, results are quite low, and the only adequate transformation seems to be the geometrical transformation originally used in \cite{ji2018iic}. This is because a transformation in the mutual information has a different effect than a transformation used for regularization. In regularization, we assume that the original image and the transformed image should contain the same class label. Thus, with KL divergence we impose that the output class probabilities of the image and the transformed should be similar. This is a way to make the loss smoother and therefore improve the optimization. In contrast, when using a transformation on the mutual information formulation does not require to maintain the same label between the original and the transformed image, it just maximizes their mutual information. Thus, in this setting stronger transformations seem to be required in order to obtain good results. This is verified by the fact that the same geometrical transformations, when used as regularization performed quite poorly (see Table \ref{tab:losses}). This is because they are too strong to be used for regularization with KL divergence. For instance on MNIST, the strong crops of the image used in these geometrical transformation could change a 9 into a 0. Thus, for mutual information new and stronger transformations than those generally used for data augmentation are needed for improved results. We leave this as future work.

Among the used transformations, we noticed that VAT performed very poorly. This is because VAT is optimised to find adversarial samples for cross-entropy. Instead, in this case, we would like to generate and use samples that are adversarial for the mutual information loss. Thus, we change the adversarial generation in equ. \ref{equ:VAT}, by using as a divergence the mutual information. In this case (IVAT) results are much more competitive.

\begin{table}[ht]
\begin{center}
\begin{small}
\begin{tabular}{l| c|c|c}
\textbf{Method} & MNIST & CIFAR10 & SVHN\\
\hline
$\MI(Y,\Y_{Geo})$ & \textbf{97.9}\ppm0.0 & \textbf{31.9}\ppm1.1 & \textbf{28.0}\ppm4.0   \\
$\MI(Y,\Y_{VAT})$ & {11.3\ppm2.2} & 17.6\ppm0.3 & 19.9\ppm0.2   \\
$\MI(Y,\Y_{IVAT})$ & {70.1\ppm7.3} & 18.8\ppm2.5 & 19.1\ppm1.9   \\
$\MI(Y,\Y_{Mixup})$ & {59.5\ppm5.6} & 20.3\ppm0.2 & 19.1\ppm1.7   \\
$\MI(Y,\Y_{Cutout})$ & {33.5\ppm1.5} & 17.1\ppm1.3 & 13.3\ppm0.3   \\
$\MI(Y,\Y_{Gauss})$ & {60.5\ppm4.4} & 17.1\ppm1.7 & 13.4\ppm-0.1  \\
\end{tabular}%
\end{small}
\end{center}
\caption{\textbf{Transformations on the mutual information}. We consider the effect of changing the transformations used in the mutual information loss report results on the three datasets validation sets. }
\label{tab:transf}
\end{table}

\begin{figure*}[ht]
\centering
\begin{subfigure}[h]{.47\linewidth}
\centering
\includegraphics[width=\textwidth]{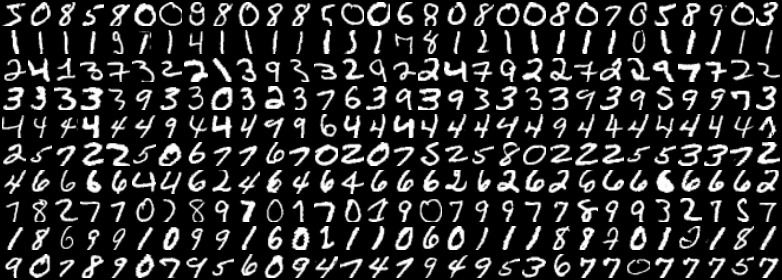}
\vspace{-.5cm}
\caption{${\MI(X,Y)}$}
\end{subfigure}
\hfill
\begin{subfigure}[h]{.47\linewidth}
\centering
\includegraphics[width=\textwidth]{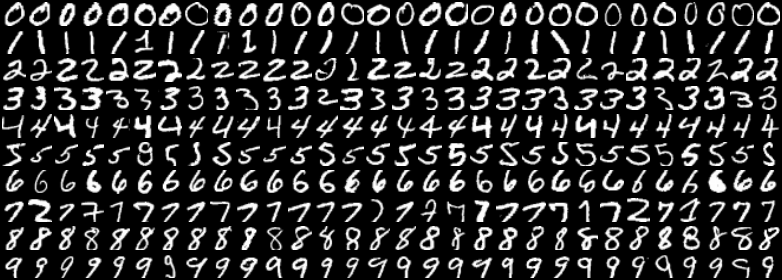}
\vspace{-.5cm}
\caption{${\MI(X,Y)}+\,D_{\mr{KL}}\big(p(Y|X),p(Y|VAT(X))\big)$}
\end{subfigure}
\vspace{0.0cm}
\centering
\begin{subfigure}[h]{.47\linewidth}
\centering
\includegraphics[width=\textwidth]{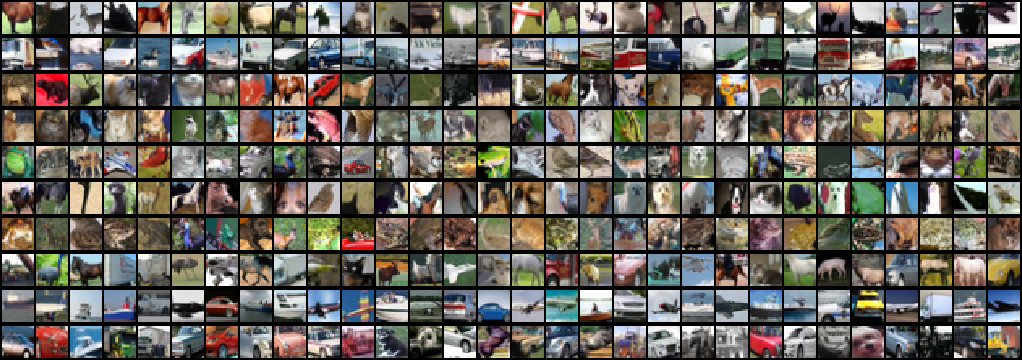}
\vspace{-.5cm}
\caption{${\MI(Y,\Y)}$}
\end{subfigure}
\hfill
\begin{subfigure}[h]{.47\linewidth}
\centering
\includegraphics[width=\textwidth]{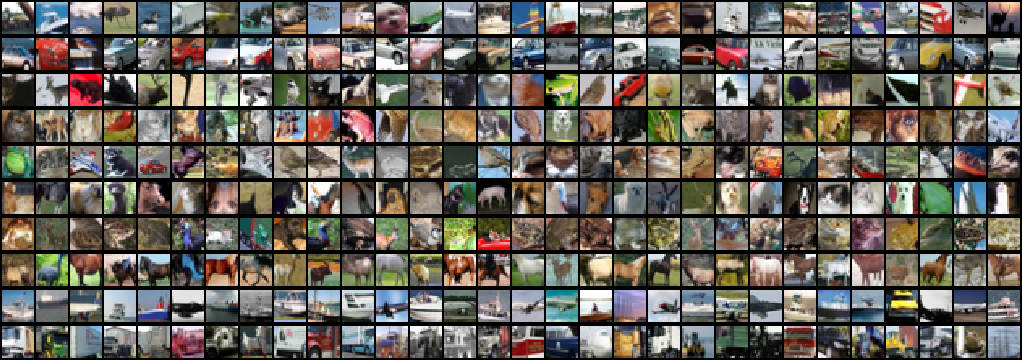}
\vspace{-.5cm}
\caption{${\MI(Y,\Y)}+\,D_{\mr{KL}}\big(p(Y|X),p(Y|Mixup(X))\big)$}
\end{subfigure}
\centering
\begin{subfigure}[h]{.47\linewidth}
\centering
\includegraphics[width=\textwidth]{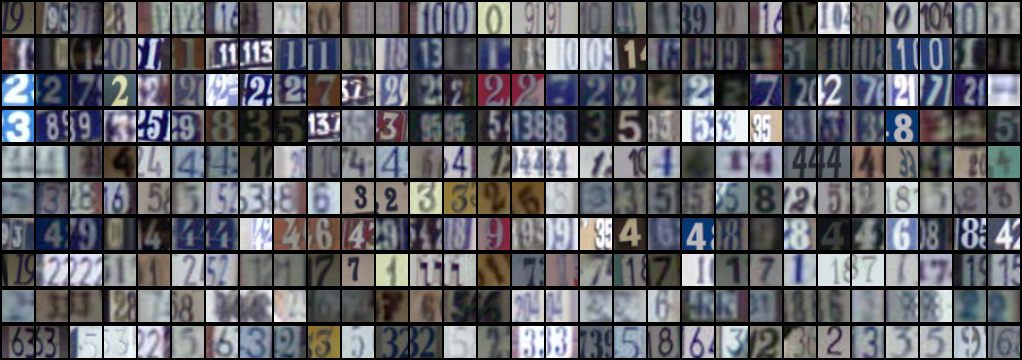}
\vspace{-.5cm}
\caption{${\MI(Y,\Y)}$}
\end{subfigure}
\hfill
\begin{subfigure}[h]{.47\linewidth}
\centering
\includegraphics[width=\textwidth]{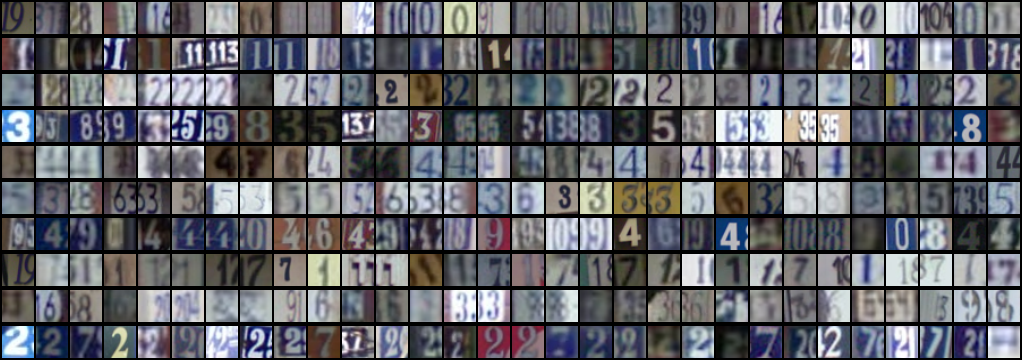}
\vspace{-.5cm}
\caption{${\MI(Y,\Y)}+\,D_{\mr{KL}}\big(p(Y|X),p(Y|Cutout(X))\big)$}
\end{subfigure}
\caption{$\bold{I(Y,\Y)}$ \textbf{with and without regularization.} We visually compare the obtained clusters for the three datasets with or without additional regularization. Each row represent a class in the dataset. Images have been randomly selected.} 
\label{fig:transform}
\end{figure*}

\begin{table}[!ht]
\begin{center}
\begin{tabular}{l|c|c}
\textbf{Training} & DA &Accuracy \\
\hline
Scratch & no & 72.9\%\\
Our Pretraining & no & \textbf{80.6}\% \\
\hline
Scratch & yes & 81.6\% \\
Our Pretraining & yes & \textbf{82.8}\% \\
\end{tabular}%
\end{center}
\caption{\textbf{Clustering for pre-training.} We evaluate how our best model can be used as initialization for supervised learning.}
\label{tab:pretraining}
\end{table}

\begin{table}[!ht]
\begin{center}
\begin{tabular}{l|c|c|c}
\textbf{Method} & FC & Conv & Y \\
\hline
VAE ~\cite{kingma2013auto} & 42.1 & 53.8 & 39.6 \\
AAE ~\cite{makhzani2015adversarial} & 43.3 & 55.2 & 37.8 \\
BiGAN ~\cite{donahue2016adversarial} & 38.4 & 56.4 & 44.9 \\
DeepInfoMax ~\cite{hjelm2018learning} & 54.1 & 63.3  & 49.6 \\
Ours   & \textbf{75.4} & \textbf{78.9} & \textbf{64.7}  \\
\end{tabular}%
\end{center}
\caption{\textbf{Supervised Linear Classification}. To compare our model with self-supervised approaches, we extract a representation from the fully connected layer (FC), the penultimate convolutional layer (Conv) and the output (Y) of our clustering network and use it for supervised training on CIFAR10.}
\label{tab:linear}
\end{table}

\subsection{Comparison with IMSAT and IIC}
In table \ref{tab:comparison_SOTA} we compare our best configuration with the results and setting presented in the IMSAT and IIC original papers on MNIST, CIFAR10 and SVHN.
The comparison with IMSAT is reported in the first three rows of the table. Our re-implementation of IMSAT correspond to $\MI(X,Y)\,+\,D_{\mr{KL}}\big(p(Y|X),p(Y|\mr{VAT}(X))\big)$ as presented in table~\ref{tab:losses}. We can see that for MNIST our values are slightly lower than the original paper. This can due to the fact that in the original implementation of IMSAT, for estimating the correct amount of noise $\epsilon$ to use, they used an adaptive formulation based on the distances between samples in pixel space. This approach works only when distances in the input space are meaningful (thus not in difficult datasets). 
For datasets other than MNIST, in the original paper \cite{hu2018imsat} clustering is not performed directly on images, but in more meaningful representations (ResNet50 feature maps pretrained on ImageNet). Instead, as our aim is to compare in fair way different methods, we use raw images on all experiments. Thus, we consider $\epsilon$ a hyper-parameter parameter.
This explains why our performance on CIFAR and SVHN are lower than in the original paper. To further verify that our implementation is competitive we also run an experiment of our IMSAT implementation with input features extracted from ResNet50 pretrained on ImageNet as in the original IMSAT paper. For this experiment on CIFAR10 we obtained an accuracy of $75.5$, which is significantly higher than the $45.6$ of the original paper.
We also compare our re-implementation of IMSAT with our best model. To make the comparison fair, we use our best model with 1 head and 1 subhead (1H1S) as in the original IMSAT paper.
For MNIST our model obtain an accuracy that is slightly inferior to IMSAT. However, for more difficult datasets (CIFAR10 and SVHN) our model clearly outperform our IMSAT baseline.

For IIC we compare the performance of our best model with the values reported in the original paper. In this case, for fair setting we use a model with 2 heads and 5 over-clustering (70 clusters) subheads (2H5S) as in the original IIC paper. In this case our model obtained an accuracy that is slightly lower than the original paper on MNIST, but significantly better on CIFAR10.



\begin{table}[ht]
\begin{center}
\begin{small}
\begin{tabular}{l|c|c|c}
\textbf{Method} & MNIST & CIFAR10 & SVHN\\
\hline
IMSAT \cite{hu2018imsat} & 98.4\ppm0.4 & 45.6\ppm0.8$^*$& 57.3\ppm3.9$^*$\\
IMSAT (our impl.) & 97.7\ppm0.3 & 18.4\ppm0.4 
& 17.1\ppm2.0\\
Our Model (1H1S) & 97.6\ppm0.1 & 37.5\ppm2.9 & 34.3\ppm3.9 \\
\hline
IIC \cite{ji2018invariant} & 98.4\ppm0.7 & 57.6\ppm5.0 & - \\
Our Model (2H5S)& 94.9\ppm2.1 & 62.5\ppm0.3 & 38.6\ppm0.8 \\

\end{tabular}%
\end{small}
\end{center}
\caption{\textbf{Comparison with IMSAT and IIC.} $^*$ means that the clustering was effectuated on higher level features from which is easier to cluster, therefore a direct comparison of the accuracies is not fair. }
\label{tab:comparison_SOTA}
\end{table}

\subsection{Clustering for pre-training}
In table~\ref{tab:pretraining}, we consider the effect of using the network trained for clustering as pre-training unsupervised initialization for a fully-supervised training on the same dataset (CIFAR10). When we train without any data augmentation, the initialization based on clustering gives an important boost in performance, going from an accuracy of $72.9\%$ for a network trained from scratch to $80.6\%$ for a network whose weights are initialized with our best model. 
When the network is trained with data augmentation, the gap between the two initialization is reduced to $1.2\%$. Thus, there is still a gain without introducing any additional data or annotations, but more limited. This is probably due to the fact that most of the useful information brought in by the clustering pre-trained model is about image transformations. Thus, when adding data augmentation that information is included directly in the training and the initialization becomes less useful.


\subsection{Disentangled Representations} 
As we perform clustering, the final results are already groups of samples that are visually similar. If the clustering works well, these groups should represent classes and they can be directly used for classification. Thus, in our experiments, it is not necessary to perform any additional supervised learning for evaluation. 
In contrast, methods based on self-supervision or that learn disentangled representations will normally need a final evaluation step in which supervision is used to learn the final classifier.
An evaluation often used for these methods consists on training the learned representation in a fully-supervised way but with a linear classifier so that its capacity is limited. 
In order to compare with methods based on self-supervision, we use our learned representation to train a linear classifier. We argue that, as clustering is very similar to the final classification task, results of our method will be better than other approaches.
Table \ref{tab:linear} compares DeepInfoMax \cite{hjelm2018learning}, a recent method for disentangled representations based on mutual information, with our clustering approach. We also report results of other popular methods from \cite{hjelm2018learning}.
We report the accuracy obtained by a linear classifier trained on the fully-connected layer (FC), penultimate convolutional layer (Conv) and output (Y), 10 values in our case and 64 for DeepInfoMax. As expected, the gap in performance is quite high: in the order of 20 points for FC, 10 for convolutional and 5 for the output Y. The reduced gap in the output is explained by the low dimensionality of the final output (10) for our model.

\section{Conclusions}
\label{sec:discussion}
In this paper, we have presented a generalization of two very popular information-based clustering approaches. We consider a clustering model based on a combination of a mutual information based loss and a regularization based on transformations. 
From our empirical evaluation we conclude that adding a regularization based on image transformations is in most of the cases beneficial. 
Our best configuration is then a clustering that maximizes the mutual information between a samples and its geometrical transformation, regularised by a KL divergence between a sample and its adversarial transformation. This configuration seems to outperform previous work on clustering as well as on disentangling representations.

\newpage

{\small
\bibliographystyle{ieee}
\bibliography{egbib}
}

\end{document}


\title{Mutual Information for Deep Clustering: an experimental study \\ Supplementary Material}


\maketitle
\ifwacvfinal\thispagestyle{empty}\fi

\begin{table}[]
\begin{small}
\begin{tabular}{lllllll}
\hline
\multicolumn{1}{c}{} & MI Var & Transform & Regularization & MNIST & CIFAR10 & SHVN \\ \hline
1 & $MI(Y, \Y)$ & Geo &  & $97.89\pm0.01\%$ & $31.86\pm1.11\%$ & $28.04\pm3.984\%$ \\
2 & $MI(Y, \Y)$ & VAT &  & $11.25\pm0.00\%$ & $17.39\pm0.39\%$ & $16.95\pm1.866\%$ \\
3 & $MI(Y, \Y)$ & Mixup &  & $53.21\pm5.62\%$ & $20.06\pm0.25\%$ & $17.13\pm1.705\%$ \\
4 & $MI(Y, \Y)$ & Geo+VAT &  & $59.39\pm41.83\%$ & $17.33\pm1.00\%$ & $15.56\pm0.675\%$ \\
5 & $MI(Y, \Y)$ & Geo+IVAT & & $93.60 \pm 7.45\%$ &	$17.64 \pm 2.60\%$	& $14.78\pm0.986\%$ \\

5 & $MI(Y, \Y)$ & Geo+mixup &  & $97.95\pm0.01\%$ & $28.14\pm0.53\%$ & $17.70\pm1.236\%$ \\
6 & $MI(Y,\Y)$ & Geo+VAT+mixup &  & $82.17\pm4.55\%$ & $18.55\pm0.82\%$ & $14.99\pm1.216\%$ \\
6 & $MI(Y,\Y)$ & Geo+IVAT+mixup &  & $91.53\pm6.78\%$ & $19.23\pm0.5\%$ & $13.66\pm0.204\%$ \\
\hilne
7 & $MI(Y, \Y)$ & Geo & VAT & $97.63\pm0.10\%$ & $36.63\pm0.56\%$ & $28.90\pm5.762\%$ \\
8 & $MI(Y, \Y)$ & Geo & Mixup & $94.07\pm6.66\%$ & $35.94\pm1.41\%$ & $25.82\pm0.620\%$ \\
9 & $MI(Y, \Y)$ & Geo & VAT+Mixup & $30.66\pm33.62\%$ & $13.60\pm6.03\%$ & $26.26\pm12.406\%$ \\
9 & $MI(Y, \Y)$ & Geo & IVAT+Mixup & $ $ & $29.81\pm1.46\%$ & $14.71\pm0.462\%$ \\
10 & $MI(Y, \Y)$ & Geo & VAT+VAT & $92.13\pm7.61\%$ & $32.24\pm4.57\%$ & $26.83\pm2.137\%$ \\
\hline
11 & $MI(X,Y)$ &  &  & $42.62\pm3.89\%$ & $16.60\pm1.02\%$ & $15.45\pm1.485$\% \\
12 & $MI(X,Y)$ &  & Geo & $43.82\pm47.8\%$ & $14.48\pm0.72\%$ & $17.42\pm0.766$\% \\
13 & $MI(X,Y)$ &  & VAT & $97.72\pm0.33\%$ & $18.36\pm0.37\%$ & $17.11\pm1.969$\% \\
14 & $MI(X,Y)$ &  & Mixup & $34.39\pm11.98\%$ & $20.11\pm1.16\%$ & $15.49\pm1.258$\% \\
15 & $MI(X,Y)$ &  & Geo+VAT & $84.47\pm11.54\%$ & $15.60\pm0.92\%$ & $19.69\pm3.108$\% \\
16 & $MI(X,Y)$ &  & Geo+Mixup & $32.37\pm10.83\%$ & $22.63\pm0.55\%$ & $14.38\pm0.563$\% \\
17 & $MI(X,Y)$ &  & VAT+Mixup & $97.23\pm0.40\%$ & $18.31\pm2.44\%$ & $17.18\pm2.107$\% \\
18 & $MI(X,Y)$ &  & Geo+VAT+Mixup & $97.57\pm0.05\%$ & $15.16\pm0.70\%$ & $16.71\pm1.995$\% \\
19 & $MI(X,Y)+MI(Y,\Y)$$ & Geo & VAT & $90.39\pm6.18\%$ & $37.45\pm12.85\%$ & $34.31\pm3.866$\% \\ \hline
\end{tabular}
\end{small}
\end{table}

{\small
\bibliographystyle{ieee}
\bibliography{egbib}
}